\documentclass[letterpaper]{article} 
\usepackage{aaai24}  
\usepackage{times}  
\usepackage{helvet}  
\usepackage{courier}  
\usepackage[hyphens]{url}  
\usepackage{graphicx} 
\usepackage{natbib}  
\usepackage{caption} 
\usepackage{caption}
\usepackage{subcaption}
\usepackage{amsfonts} 
\usepackage{amsmath}
\usepackage{multirow}
\usepackage{booktabs}
\usepackage{algorithm}
\usepackage{algorithmic}

\usepackage{newfloat}
\usepackage{listings}
\DeclareCaptionStyle{ruled}{labelfont=normalfont,labelsep=colon,strut=off} 
\floatstyle{ruled}
\newfloat{listing}{tb}{lst}{}
\floatname{listing}{Listing}
\title{Large Language Models Are Clinical Reasoners:\\ Reasoning-Aware Diagnosis Framework with Prompt-Generated Rationales}
\author{
    Taeyoon Kwon\textsuperscript{\rm 1}\equalcontrib,
    Kai Tzu-iunn Ong\textsuperscript{\rm 1}\equalcontrib,
    Dongjin Kang\textsuperscript{\rm 2},
    Seungjun Moon\textsuperscript{\rm 2},
    Jeong Ryong Lee\textsuperscript{\rm 3}, \\
    Dosik Hwang\textsuperscript{\rm 3},
    Beomseok Sohn\textsuperscript{\rm 4},
    Yongsik Sim\textsuperscript{\rm 4},
    Dongha Lee\textsuperscript{\rm 1},
    Jinyoung Yeo\textsuperscript{\rm 1}
}
\affiliations{
    \textsuperscript{\rm 1}Department of Artificial Intelligence, Yonsei University \\
    \textsuperscript{\rm 2}Department of Computer Science and Engineering, Yonsei University \\
    \textsuperscript{\rm 3}Department of  Electrical and Electronic Engineering, Yonsei University \\
    \textsuperscript{\rm 4}Department of Radiology, College of Medicine, Yonsei University

 }

\newcommand{\eg}[0]{\textit{e.g.}}
\newcommand{\ie}[0]{\textit{i.e.}}

\begin{document}

\maketitle

\begin{abstract}

Machine reasoning has made great progress in recent years owing to large language models (LLMs). In the clinical domain, however, most NLP-driven projects mainly focus on clinical classification or reading comprehension, and under-explore clinical reasoning for disease diagnosis due to the expensive rationale annotation with clinicians. In this work, we present a ``reasoning-aware'' diagnosis framework that rationalizes the diagnostic process via prompt-based learning in a time- and labor-efficient manner, and learns to reason over the prompt-generated rationales. Specifically, we address the clinical reasoning for disease diagnosis, where the LLM generates diagnostic rationales providing its insight on presented patient data and the reasoning path towards the diagnosis, namely \textbf{Clinical Chain-of-Thought (Clinical CoT)}. We empirically demonstrate LLMs/LMs' ability of clinical reasoning via extensive experiments and analyses on both rationale generation and disease diagnosis in various settings. 
We further propose a novel set of criteria for evaluating machine-generated rationales' potential for real-world clinical settings, facilitating and benefiting future research in this area.\footnote{We release the few-shot exemplars for Clinical Chain-of-Thought at \url{https://github.com/ktio89/ClinicalCoT}}

\end{abstract}

\section{Introduction}
\label{sec:introduction}

Reasoning is the ability to assess things logically based on available information of various types. Reasoning in clinical diagnosis, also known as clinical reasoning or diagnostic reasoning, is a dynamic thinking process between the observed clinical evidence and the identification of disease. 
It involves an integration of patient data, relevant medical knowledge, clinicians' experience, and other contextual or situational factors~\citep{norman2005research, cook2018management}.
Poor clinical reasoning has been directly linked to misdiagnoses and eventually causing hospital adverse events including patient death~\citep{balogh2015improving}.
Therefore, effective clinical reasoning is crucial for diagnosis in real clinical settings~\citep{kassirer1989diagnostic}.

Recently, deep learning (DL) models are widely utilized for disease diagnosis. However, a predominant portion of existing approaches formulates the process simply as image or text classification~\citep{bakator2018deep, kumar2022artificial}. 
These approaches entirely exclude the aforementioned clinical reasoning in their modeling and focus on fine-tuning high-capacity models for better feature extraction~\citep{jang2022m3t}. However, such a data-driven approach can be limited by the data-scarcity problem in biomedical domains.
Moreover, high-capacity models are shown to memorize the dataset, rather than solving the diagnosis task through logical reasoning~\citep{mitra2019exploring}, and they cannot provide explanations justifying their diagnoses. 
Since whether a diagnosis can be explained and whether it matches the reasoning of humans are important for gaining clinicians' trust in DL techniques~\citep{holzinger2017we}, this naive approach largely limits models' potential to be implemented for real-world applications.

Meanwhile, large language models have demonstrated their ability to perform multi-step reasoning as well as present the thinking process behind it, which is known as chain-of-thought (CoT) reasoning~\citep{wei2022chain}. Previous works have applied such reasoning ability to various domains~\citep{wei2022chain, kojima2022large, wu2023chain, lievin2022can}. In these works, LLMs serve as reasoners that generate natural language rationales guiding and explaining the solution.
Despite such success, the use of LLMs to address clinical reasoning in disease diagnosis for real-world applications is still an under-explored area at the moment.

\begin{figure*}[t!]
\centering
\includegraphics[width=0.95\textwidth]{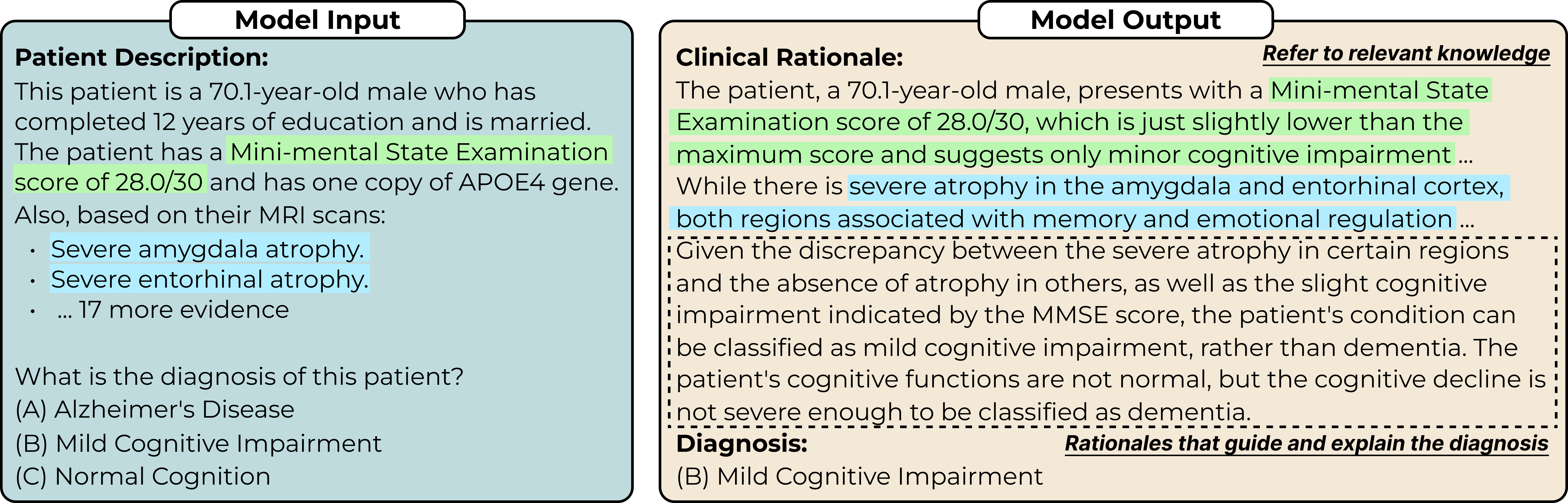} 
\caption{Clinical reasoning in disease diagnosis.}
\label{fig:model_io}
\end{figure*}

Motivated by these, in this work, we make a step toward clinical reasoning in disease diagnosis, where the models are aware of the clinical reasoning behind the diagnosis, as illustrated in Figure~\ref{fig:model_io}.
To this end, we formulate the clinical reasoning in disease diagnosis as chain-of-thought reasoning, namely Clinical Chain-of-Thought (Clinical CoT).
Our goal is to facilitate clinical reasoning by leveraging LLMs to reason over patient data, refer to relevant knowledge, and generate rationales that guide and explain the diagnosis.

Our contributions are two-fold: (i) We propose a practical framework for reasoning-aware diagnosis. 
Our framework involves clinical rationalization that augments the existing clinical data with clinical rationales, few-shot reasoning and diagnosis with LLMs, and distillation towards smaller models.
(ii) We conduct a thorough analysis of the framework in our testbed diagnosis dataset to gain a deep understanding of the clinical reasoning task. We show that by reasoning over presented clinical data, models can achieve better performance in disease diagnosis. Also, our extensive evaluation and analysis of generated rationales demonstrate that both the LLMs and distilled models can replicate the reasoning of clinical professionals in a human-like manner.

\section{Problem Formulation}
\label{sec:problem_formulation}
\subsection{Clinical Reasoning for Disease Diagnosis}
Most existing approaches for diagnosing diseases with DL models formulate the process simply as image or text classification~\citep{bakator2018deep, kumar2022artificial}.
That is, given patient description $\mathcal{P}$, such as medical images or electronic health records, a diagnosis model $\theta$ is trained to predict the correct diagnosis $\mathcal{D}$:
\begin{equation}
\label{eq:disease_diagnosis}
    \mathcal{D} \sim P_{\theta}(\cdot|\mathcal{P})
\end{equation}
However, this approach neglects the clinical reasoning connecting the presented patient description and the final diagnosis~\citep{kassirer1989diagnostic}. 
The absence of effective clinical reasoning can lead to diagnostic errors (\eg, misdiagnoses), which are reported to contribute to around 10\% of patient deaths and hospital adverse events~\citep{norman2005research}.

To address that, we exploit LLMs' reasoning capacity in clinical diagnosis, where the LLMs ought to perform clinical reasoning over presented clinical data. 
Formally, given patient description $\mathcal{P}$, the model first generates a rationale $\mathcal{R}$ decomposing the reasoning process over $\mathcal{P}$, and then makes its diagnosis $\mathcal{D}$ based on $\mathcal{P}$ and $\mathcal{R}$:
\begin{align}
\label{eq:reasoning_diagnosis}
    \mathcal{R} \sim P_{\theta}(\cdot|\mathcal{P}) \\
    \mathcal{D} \sim P_{\theta}(\cdot|\mathcal{P}, \mathcal{R})
\end{align}

\subsection{Testbed: Alzheimer's Disease Diagnosis}

Alzheimer's disease (AD) is an irreversible neurodegenerative disease associated with cognitive decline~\citep{deture2019neuropathological}.
In this study, we choose AD diagnosis task as the testbed for clinical reasoning. 
This choice is based on the fact that AD diagnosis requires a thorough understanding of various aspects of the disease~\citep{budson2012new}. 
In our study, patient description $\mathcal{P}$ consists of (1) textual descriptions derived from the MRI scan, such as \textit{``This patient has \textbf{SEVERE} hippocampal atrophy''};\footnote{Hippocampal atrophy refers to the shrinkage or loss of nerve cells in the hippocampus, a region related to memory formation and memory retrieval~\citep{voss2017closer}.} (2) demographic information; (3) educational level; (4) results from the mini-mental state examination (MMSE); (5) the presence of APOE4 allele. 
The diagnosis $\mathcal{D}$ can be either \textit{Alzheimer's Disease}, \textit{Mild Cognitive Impairment} (MCI), or \textit{Normal Cognition} (NC).
Details on transforming MRI scans into textual descriptions are provided in the appendix.

\section{Reasoning-Aware Diagnosis Framework}

\begin{figure*}[t!]
\centering
\includegraphics[width=0.9\textwidth]{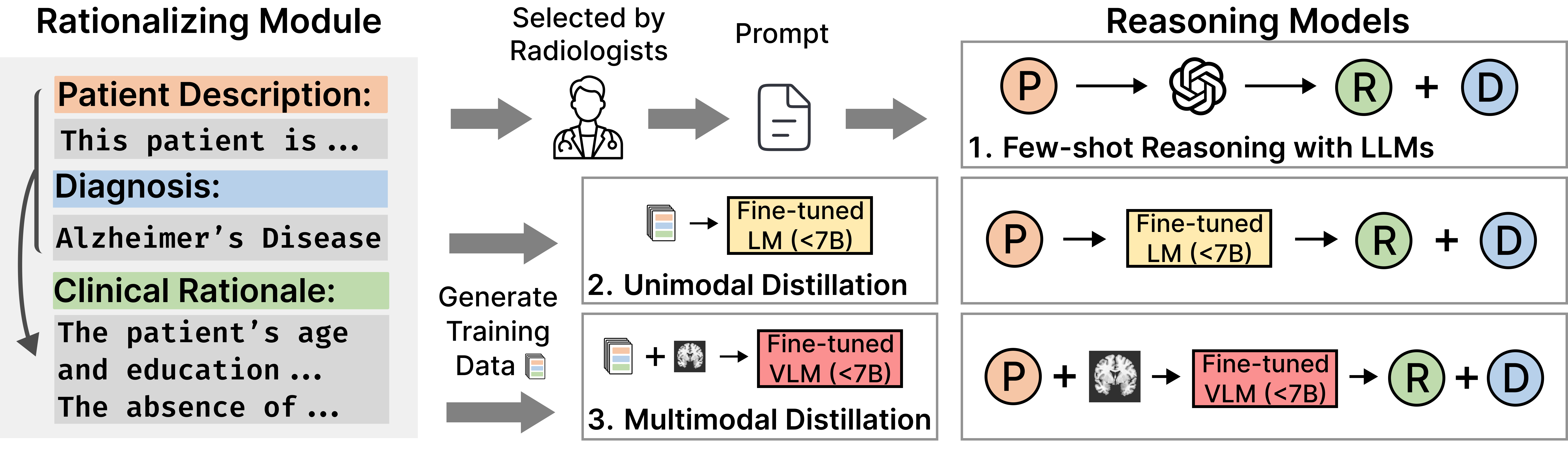} 
\caption{An overview of our framework ($\mathcal{P}$: Patient description; $\mathcal{D}$: Diagnosis; $\mathcal{R}$: Clinical rationale).}
\label{fig:framework}
\end{figure*}

\subsection{Framework Overview}
Recent works successfully leverage LLMs' ability of CoT reasoning to generate free-text rationales that present the reasoning path and the necessary knowledge towards the answers in various reasoning tasks~\citep{wei2022chain, wu2023chain}. 
Our goal is to exploit such ability in clinical diagnosis, where the LLMs ought to generate rationales demonstrating its reasoning over presented clinical data. For that, we formulate the rationale generation in clinical diagnosis as Clinical CoT reasoning.
Upon that, we propose a reasoning-aware diagnosis framework (Figure~\ref{fig:framework}), which includes modules addressing different approaches to facilitate clinical reasoning.

\subsection{Module I: Clinical Rationalization}
\label{ssec:rationalizing}
To generate clinical CoT rationales, which deliver the diagnostic reasoning towards the correct diagnosis, by prompting an LLM to rationalize the presented clinical data.

Formally, given clinical data consisting of patient description $\mathcal{P}$ and a ground-truth label of the diagnosis $\mathcal{D}$, which can be either \textit{Alzhemer's Disease}, \textit{Mild Cognitive Impairment} (MCI), or \textit{Normal Cognition} (NC), the LLM is prompted to generate clinical rationales $\mathcal{R}^*$ that demonstrate the reasoning process over $\mathcal{P}$ such that the final diagnosis $\mathcal{D}$ can be induced from $\mathcal{R}^*$:
\begin{equation}
\label{eq:extract_cot_1}
    \mathcal{R}^* = \underset{\mathcal{R}}{\text{argmax}} \, P_{\text{LLM}}(\mathcal{R}| \mathcal{P}, \mathcal{D})
\end{equation}

As output, we collect a set $\mathbb{D}$ of the processed samples, each of which is a triplet of patient description $\mathcal{P}$, a ground-truth label of the diagnosis $\mathcal{D}$, and the clinical rationales $\mathcal{R}$.

\subsection{Module II-1: Few-shot CoT Reasoning}
\label{ssec:few_shot_diagnosis}

LLMs have demonstrated promising performance in tasks requiring logical reasoning with CoT prompting~\citep{kojima2022large, wei2022chain}.
As a pioneer study towards clinical reasoning with LLMs, we investigate if such success can be replicated in the domain of clinical diagnosis.
Thereby, our second module addresses few-shot disease diagnosis, where we prompt LLMs to perform clinical reasoning before the diagnosis. 

Formally, given the patient description $\mathcal{P}$, an LLM is prompted to generate both a plausible clinical rationale $\hat{\mathcal{R}}$ and the name of the predicted diagnosis $\hat{\mathcal{D}}$:
\begin{align}
\label{eq:extract_cot_2}
    \hat{\mathcal{R}} = \underset{\mathcal{R}}{\text{argmax}}\,  P_{\text{LLM}}(\mathcal{R}|\mathcal{P}) \\
    \Rightarrow \, \hat{\mathcal{D}} = \underset{\mathcal{D}}{\text{argmax}}\,
    P_{\text{LLM}}(\mathcal{D}|\mathcal{P}, \hat{\mathcal{R}})
\end{align}
where $\Rightarrow$ indicates a sequential generation of tokens.

\begin{table*}[t!]
\centering
\setlength{\tabcolsep}{4pt}
\small
\begin{tabular}{lcccccccc|ccccccc}
\toprule
      &         & \multicolumn{7}{c}{\textbf{ADNI}} & \multicolumn{7}{c}{\textbf{AIBL}}  \\ \cmidrule(lr){3-9} \cmidrule(lr){10-16}
      &         &  Accuracy  & \multicolumn{3}{c}{Precision} & \multicolumn{3}{c}{Recall} & Accuracy & \multicolumn{3}{c}{Precision} & \multicolumn{3}{c}{Recall}\\ \cmidrule(lr){3-3} \cmidrule(lr){4-6} \cmidrule(lr){7-9} \cmidrule(lr){10-10} \cmidrule(lr){11-13} \cmidrule(lr){14-16}
Model & Prompt & Total & AD & MCI & NC  & AD & MCI & NC & Total & AD & MCI & NC  & AD & MCI & NC \\ 
\midrule
ChatGPT & 0-shot  & 55.3 & 56.8 & 45.3 & 69.9 & 96.4 & 22.4 & 48.8 & 54.0 & 59.8 & 47.2 & 61.1 & 80.0 & 31.7 & 55.0 \\
        & 1-shot  & 50.7 & 61.4 & 40.6 & 71.4 & 81.5 & \textbf{62.9} &  7.9 & 50.7 & 67.4 & 41.8 & \textbf{70.6} & 66.9 & \textbf{74.7} &  8.6 \\
        & 3-shot  & 58.8 & 61.6 & 47.1 & \textbf{71.5} & 85.5 & 46.7 & 44.8 & 57.2 & \textbf{68.3} & 46.6 & 63.5 & 63.1 & 57.0 & 52.1 \\
        & 5-shot  & 57.3 & 55.7 & 44.4 & 70.2 & \textbf{97.2} & 23.2 & 53.2 & 56.3 & 57.7 & \textbf{47.3} & 62.5 & \textbf{86.9} & 33.5 & 53.6 \\
        & Clinical CoT  & \textbf{67.3} & \textbf{62.2} & \textbf{54.2} & 62.7 & 90.3 & 26.6 & \textbf{86.5} & \textbf{62.4} & 63.8 & 43.7 & 53.2 & 79.2 & 25.3 & \textbf{88.6}  \\
\midrule
GPT-4   & 0-shot  & 59.6 & 51.1 & 51.6 & 76.8 & \textbf{99.6} & 24.3 & 42.1 & 55.4 & 52.1 & 49.5 & \textbf{71.1} & 93.9 & 29.1 & 49.3 \\
        & 1-shot  & 53.0 & 54.1 & 44.4 & \textbf{81.0} & 98.4 & 42.9 & 18.7 & 54.9 & 55.6 & 42.2 & 65.5 & \textbf{96.2} & 38.0 & 35.7 \\
        & 3-shot  & 61.8 & 64.3 & 50.9 & 76.5 & 86.3 & 55.2 & 44.4 & 58.2 & 66.2 & 46.7 & 67.0 & 72.3 & 53.2 & 50.7 \\
        & 5-shot  & 62.6 & 67.8 & 50.5 & 76.5 & 90.8 & \textbf{57.5} & 40.1 & 59.8 & 68.9 & 48.4 & 69.3 & 78.5 & \textbf{58.9} & 43.6 \\
        & Clinical CoT  & \textbf{68.4} & \textbf{77.5} & \textbf{59.3} & 67.4 & 76.2 & 40.5 & \textbf{89.3} & \textbf{62.6} & \textbf{82.2} & \textbf{51.2} & 60.9 & 63.8 & 39.2 & \textbf{87.9} \\
\bottomrule
\end{tabular}%
\caption{Evaluation on LLMs in zero-and-few-shot diagnosis. The Clinical CoT includes two exemplar shots.}
\label{tab:fewshot}
\end{table*}

\subsection{Module II-2: Unimodal-Student Distillation}
Despite the impressive performance offered by LLMs in few-shot settings, it is non-trivial to deploy them for real-world applications due to the large size of parameters.\footnote{One LLM with 175B parameters requires at least 350GB GPU memory with tailored infrastructures~\citep{zheng2022alpa}.} Recent works resolve this by using LLMs to augment the target dataset with rationales and use the augmented dataset to fine-tune smaller models, aiming to distill LLMs' reasoning capacity into models that are more affordable for practical uses~\citep{hsieh-etal-2023-distilling}. This approach is known as knowledge distillation~\citep{hinton2015distilling}.

This module distills the knowledge of diagnostic reasoning from the LLM (teacher) into orders-of-magnitude smaller language models, with the goal of developing smaller CoT reasoners for real clinical settings.
Applying our rationalizing module (Module I), we obtain clinical data for AD diagnosis that are augmented with clinical rationales. We purpose this augmented dataset as training data to train the student language models.

Formally, given patient description $\mathcal{P}$, the LM is trained to sequentially predict the clinical rationale $\mathcal{R}$ and the ground-truth label of the diagnosis $\mathcal{D}$. The language model is optimized by minimizing the generation loss $\cal{L}_{\text{LM-Distill}}$:
\begin{equation}
    \mathcal{L}_{\text{LM-Distill}} = \underset{(\mathcal{P},\mathcal{D},\mathcal{R}) \in \mathbb{D}}{\mathbb{E}} [- \log P_{\text{LM}}(\mathcal{R}, \mathcal{D} | \mathcal{P})] 
\end{equation}

\subsection{Module II-3: Multimodal-Student Distillation}
Besides training smaller CoT reasoners with language models, multimodal CoT, where both visual and textual inputs can be considered via vision-language models (VLMs), has also garnered attention.
For instance, \citet{zhang2023multimodal} showed that by including images alongside textual inputs, models with under 1B parameters can generate more effective CoT rationales and vastly outperform the previous state-of-the-art LLM with 175B parameters on a question-answering
benchmark. 
Meanwhile, the diagnosis of many diseases including AD involves medical images such as MRI scans, fundus photographs, and X-ray images~\citep{kumar2022artificial}. 
Therefore, we further extend knowledge distillation in clinical diagnosis to VLMs.

Formally, given patient description $\mathcal{P}$ and its corresponding MRI scan $\mathcal{V}$, the VLM is trained to sequentially predict the clinical CoT rationale $\mathcal{R}$ and the ground-truth label of the diagnosis $\mathcal{D}$ based on $\mathcal{P}$ and $\mathcal{V}$. The VLM is learnt by minimizing the generation loss $\cal{L}_{\text{VLM-Distill}}$:
\begin{equation}
    \mathcal{L}_{\text{VLM-Distill}} = \underset{(\mathcal{P},\mathcal{V},\mathcal{D},\mathcal{R}) \in \mathbb{D}}{\mathbb{E}} [- \log P_{\text{VLM}}( \mathcal{R}, \mathcal{D} | \mathcal{P}, \mathcal{V})]
\end{equation}

\section{Experiments}
\subsection{Experimental Settings}
\paragraph{Dataset.}
We acquire 7,124 clinical data for Alzheimer's disease (AD) from the Alzheimer’s Disease Neuroimaging Initiative (ADNI)~\citep{jack2008alzheimer} and 428 from Australian Imaging Biomarkers and Lifestyle Study of Ageing (AIBL)~\citep{ellis2009australian}.
Datasets from these organizations have profoundly facilitated the study of AD. 
Both datasets comprise three components: (1) MRI scans, (2) ground-truth labels of diagnosis, and (3) patient descriptions, including demographic information, educational level, results from the mini-mental state examination (MMSE) and the presence of APRO4 allele.
Data from ADNI are split into training, validation, and test sets.
All the AIBL data are exclusively used as an out-of-domain test set rather than training. The ratio of AD:MCI:NC in both test sets is roughly 1:1:1.

\begin{figure}[t!]
\centering
\includegraphics[width=\columnwidth]{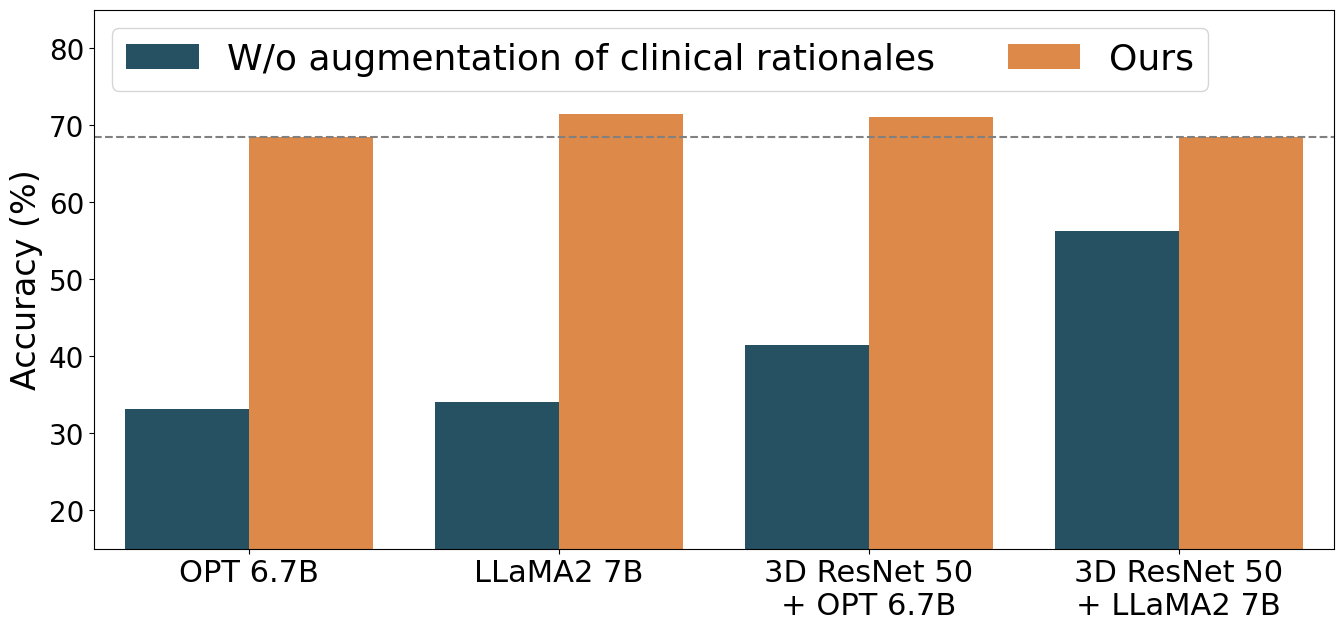} 
\caption{Performance of student models trained with and without clinical rationales, reported on ADNI. The dotted line is the performance of the teacher LLM (GPT-4).}
\label{fig:ablation_with_or_without}
\end{figure}

\paragraph{Large language models.}
We choose ChatGPT~\citep{openai2023chatgpt} and GPT-4~\citep{openai2023gpt4} as our selection for LLMs. They have shown an impressive ability to perform chain-of-thought reasoning. For our clinical rationalization module (Module I), we adopt GPT-4. And we adopt both ChatGPT and GPT-4 for few-shot diagnosis (Module II).
In all of our experiments, we set the temperature to 0.7, max tokens to 2000, and apply greedy decoding.

\paragraph{Unimodal-student models.}
We consider OPT~\citep{zhang2022opt} and LLaMA2~\citep{touvron2023llama2}, two commonly used language models as our foundation model for the unimodal student.
For our experiments, we use the 1.3B and 6.7B versions of OPT and the 7B version of LLaMA2.

\paragraph{Multimodal-student models.}
Following \citet{tsimpoukelli2021multimodal}, our multimodal student is based on the vision-language model, which consists of convolutional neural networks as vision encoder and a language model as text encoder. The vision encoder extracts image features from the MRI scan. During training, the vision encoder is trained to align its extracted feature with text features, such that the language model can effectively attend to features of both modalities.
Since MRI scans are 3-dimensioanl images, we consider 3D ResNet~\citep{hara2017learning} as our foundation model of vision encoder. 
For the language model, we use the 1.3B and 6.7B versions of OPT and the 7B version of LLaMA2.
Implementation details of our student models are provided in the appendix.

\begin{table*}[ht!]
\centering
\setlength{\tabcolsep}{4pt}
\small
\begin{tabular}{lccccccc|ccccccc}
\toprule
      & \multicolumn{7}{c}{\textbf{ADNI} (In-domain)} & \multicolumn{7}{c}{\textbf{AIBL} (Out-of-domain)}  \\ \cmidrule(lr){2-8} \cmidrule(lr){9-15}
      &    Accuracy   & \multicolumn{3}{c}{Precision} & \multicolumn{3}{c}{Recall} &  Accuracy   & \multicolumn{3}{c}{Precision} & \multicolumn{3}{c}{Recall}\\  \cmidrule(lr){2-2} \cmidrule(lr){3-5} \cmidrule(lr){6-8} \cmidrule(lr){9-9} \cmidrule(lr){10-12} \cmidrule(lr){13-15}
Baselines & Total & AD & MCI & NC  & AD & MCI & NC & Total & AD & MCI & NC  & AD & MCI & NC \\ 
\midrule
GPT-4 (Teacher Model) & 68.4 & 77.5 & 59.3 & 67.4 & 76.2 & 40.5 & 89.3 & 62.6 & 82.2 & 51.2 & 60.9 & 63.8 & 39.2 & 87.9 \\
3D ResNet-50         & 49.8 & 78.0 & 37.9 & 59.1 & 44.3 & 67.1 & 37.3 & 48.1 & 80.5 & 40.1 & 50.7 & 63.8 & 39.2 & 87.9 \\
3D ResNet-152        & 51.9 & 77.0 & 37.1 & 52.4 & 55.6 & 45.1 & 55.1 & 47.6 & 61.1 & 39.8 & 61.1 & 33.8 & 68.3 & 37.1 \\
\midrule
\multicolumn{15}{l}{Unimodal Students} \\
\midrule
OPT 1.3B          & 66.5 & 78.9 & 52.3 & \textbf{73.3} & 65.5 & \textbf{61.8} & 72.6 & \textbf{70.0} & \textbf{88.4} & 56.9 & \textbf{80.2} & 76.2 & \textbf{81.0} & 52.1 \\
OPT 6.7B          & 68.4 & 77.4 & 61.3 & 64.9 & \textbf{85.5} & 37.8 & 83.7 & 66.1 & 79.0 & 59.6 & 60.2 & \textbf{83.8} & 37.3 & 82.1 \\
LLaMA2 7B         & \textbf{71.5} & \textbf{83.2} & \textbf{63.0} & 67.9 & 81.9 & 48.6 & \textbf{84.9} & 69.4 & 84.9 & \textbf{60.0} & 65.8 & 82.3 & 57.0 & \textbf{71.4} \\
\midrule
\multicolumn{15}{l}{Multimodal Students} \\ 
\midrule
\multicolumn{15}{l}{3D ResNet-50 (0.05B)} \\
+ OPT 1.3B          & 68.6 & 86.1 & 53.3 & 76.4 & 75.4 & 71.4 & 59.1 & 65.6 & 87.3 & 52.9 & 70.7 & 69.2 & 73.4 & 53.5 \\
+ OPT 6.7B          & 70.8 & \textbf{89.5} & \textbf{56.0} & 74.8 & 75.8 & 70.7 & 65.9 & 65.7 & 87.4 & 53.5 & 67.2 & 69.2 & 67.7 & 60.0 \\
+ LLaMA2 7B         & 69.0 & 82.8 & \textbf{56.0} & 69.3 & 83.5 & 57.9 & 66.3 & 68.0 & 84.8 & 55.7 & \textbf{73.9} & \textbf{81.5} & \textbf{74.1} & 48.6 \\ \cmidrule(lr){1-15}
\multicolumn{15}{l}{3D ResNet-152 (0.1B)} \\
+ OPT 1.3B          & 68.9 & 79.6 & 54.8 & 75.3 & 79.0 & 61.3 & \textbf{66.6} & 67.2 & 73.8 & \textbf{66.4} & 62.1 & 79.3 & 55.8 & \textbf{73.1} \\
+ OPT 6.7B          & \textbf{71.0} & 88.9 & 55.7 & \textbf{79.5} & 74.2 & \textbf{75.7} & 63.0 & 65.6 & \textbf{88.0} & 53.0 & 66.1 & 74.4 & 67.1 & 55.7 \\
+ LLaMA2 7B         & 68.5 & 83.3 & 54.9 & 69.2 & \textbf{84.7} & 60.6 & 60.7 & \textbf{69.4} & 85.5 & 57.9 & 72.0 & \textbf{81.5} & 72.2 & 55.0 \\
\bottomrule
\end{tabular}
\caption{Evaluation on unimodal and multimodal students compared with GPT-4 and vision-only baselines in AD diagnosis. The GPT-4 (Teacher Model) refers to the performance of GPT-4 augmented with clinical CoT rationales in Table~\ref{tab:fewshot}.
}
\label{tab:student}
\end{table*}

\subsection{Results and Discussion}
We now present the empirical findings of the following
research questions that guide our experiments:
\newline\textbf{RQ1:} \textit{Does clinical rationales improve AD diagnosis?} 
\newline\textbf{RQ2:} \textit{Does knowledge distillation benefit small models?}
\newline\textbf{RQ3:} \textit{Is our framework helpful in data-scarce scenarios?}
\newline\textbf{RQ4:} \textit{What causes misdiagnosis and how to get over it?}

\paragraph{LLMs' diagnostic performance (RQ1).} Table~\ref{tab:fewshot} presents the experimental results. 
Firstly, we observe that LLMs without clinical CoT generally yield high recall in one class (mostly AD) with fairly low recall in the other two classes.
This suggests that clinical CoT can prevent LLMs from being biased toward a specific diagnosis.

Secondly, LLMs with clinical CoT show huge improvements in accuracy compared to baselines with more shots. This follows the observation in \citet{kojima2022large}, where LLMs with 2-shot CoT prompting largely outperform LLMs with 8-shot standard prompting in arithmetic domains. We demonstrate the same patterns in clinical diagnosis.

\paragraph{Performance of student models (RQ2).} 
Table~\ref{tab:student} presents the experimental results of student models.
Firstly, among unimodal students: For in-domain data, 
OPT 6.7B shows comparable performance to the teacher model in total accuracy. 
LLaMA2 7B yields a significant performance gain in accuracy and precision in all three classes;
For out-of-domain data, all unimodal students outperform the teacher LLM in accuracy. In addition, OPT 1.3B and LLaMA2 7B demonstrate better precision for all classes and higher recall for AD and MCI.
Moreover, all the text-only unimodal students outperform the baseline vision models in total accuracy, despite the image-intensive nature of AD diagnosis.

Secondly, for multimodal students: Similar to the finding from \citet{zhang2023multimodal}, with multimodal CoT, all VLMs with orders-of-magnitude smaller sizes of parameters outperform the LLM teacher in total accuracy as well as recall for AD and/or MCI. 
Also, all multimodal students achieve remarkably higher total accuracy than vision-only baselines.

\begin{figure}[t!]
\centering
\includegraphics[width=1\columnwidth]{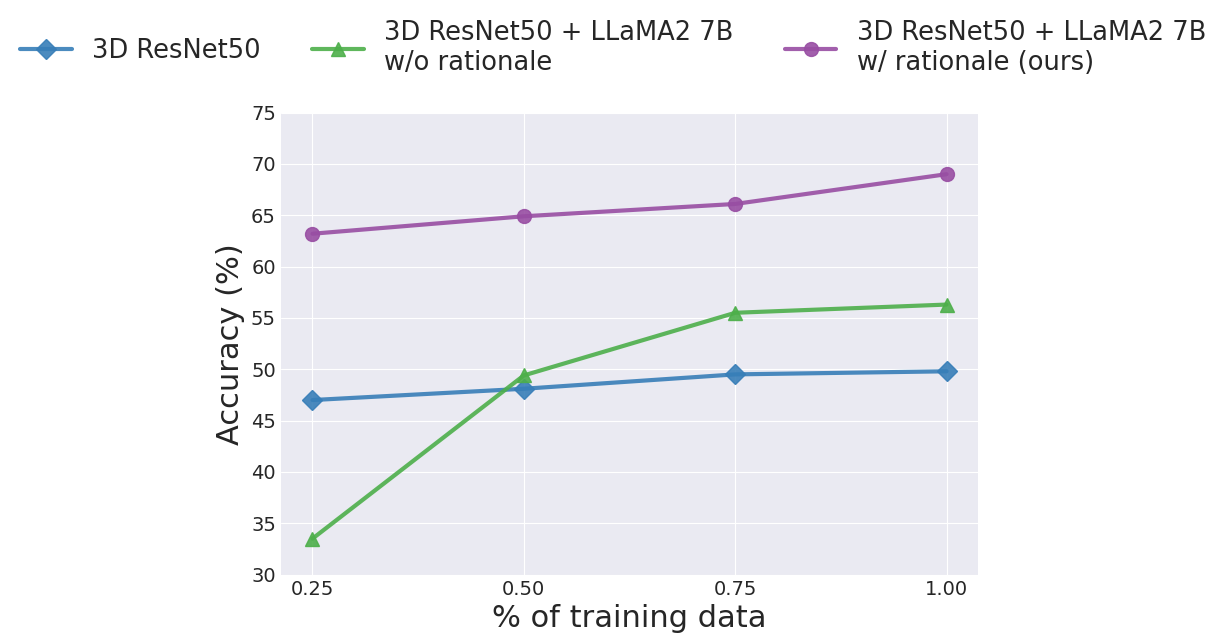} 
\caption{Data efficiency brought by clinical reasoning.}
\label{fig:ablation_data_efficiency}
\end{figure}

Overall, although metrics in which student models win slightly differ when the adopted encoders change, we can conclude a general observation: most of the students exhibit higher accuracy and recall for AD and MCI than the LLM teacher.
A higher recall indicates that the approach is more effective at minimizing false negatives, which suggests that the student models are better at avoiding misdiagnosing patients with AD or MCI as normal.

Figure~\ref{fig:ablation_with_or_without} visualizes the difference in diagnostic accuracy of our student models with and without knowledge distillation from the LLM. We can clearly observe the significant improvement in performance when training the model with our data augmented with clinical rationales.

\paragraph{Data efficiency (RQ3).}
The lack of sufficient data is a long-standing problem in the biomedical domain. 
Thus, we experiment on our multimodal student with varying amounts of training data to examine if our framework is helpful in data-scarce scenarios. The results are in Figure~\ref{fig:ablation_data_efficiency}.

Our multimodal student (purple), trained with clinical rationales, consistently outperforms the vision-only and vision-language baseline models no matter how much training data is used.
Furthermore, when trained with only 25\% of training data, the multimodal student exhibits higher accuracy than both baselines trained with 100\% of data. 
Our approaches show comparable performance even when only a limited amount of training data is available.
These experimental findings confirm the data efficiency brought by distilling LLMs' reasoning capacity to small diagnosis models, which is an important property in the biomedical domain.

\begin{figure}[t!]
\centering
\includegraphics[width=1\columnwidth]{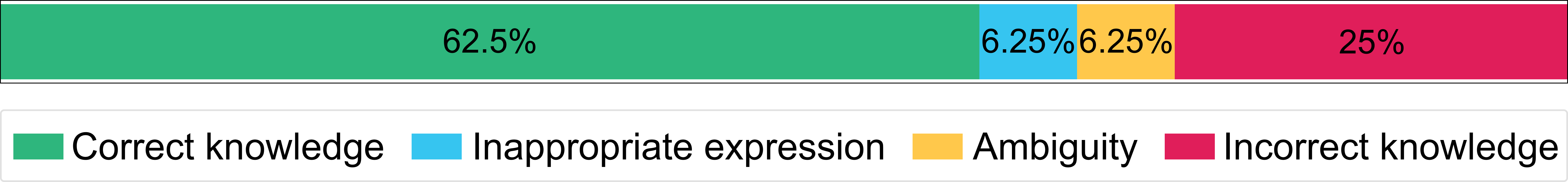} 
\caption{Analysis of rationales from GPT-4's misdiagnoses.}
\label{fig:failed_cases}
\end{figure}

\paragraph{Analysis of misdiagnosed cases (RQ4).}
\label{ssec:ineffective}
Since the rationales are generated ``during'' the diagnosis, investigating the correlation between generated rationales and misdiagnoses is rather important.
For that, we sample 32 rationales generated from misdiagnosed cases of GPT-4 and present the analysis in Figure~\ref{fig:failed_cases} (by two radiologists).


\begin{figure*}[th!]
\centering
\includegraphics[width=\textwidth]{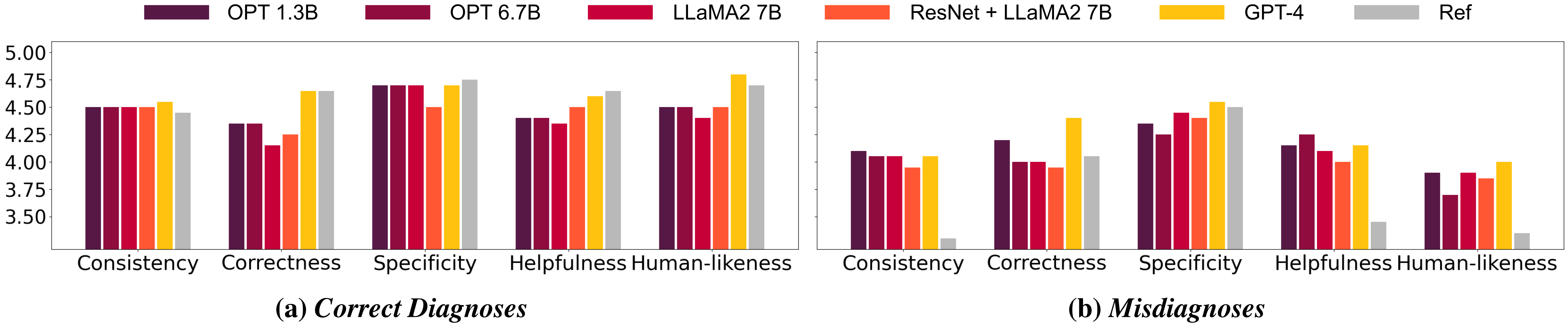}
\caption{Evaluations on clinical rationales. We report the average score (score range: 0-5).}
\label{fig:human_eval}
\end{figure*}

In failed cases, 75\% of the rationales' contents are all medically correct. Among them, some contain expressions that generally will not be used in radiology reports, such as \textit{``Interestingly, ...''} (Inappropriate expression; 6.25\%) or show wrong usages of language due to the discrepancy between clinical and general domains (Ambiguity; 6.25\%). For example, a rationale uses \textit{``positive sign''} in a scenario that is ``good for the patient''. However, this term is often used in the context of ``abnormal''.

Only 25\% of them contain at least one medically incorrect knowledge;  This shows that \textit{the misdiagnosis is not necessarily caused by ineffective rationales}. We presume optimizing the modeling on how to utilize rationales for the diagnosis (\eg, dividing the generation of rationales and the diagnosis into two separate stages) may possibly lead to better performance. We leave this to future works.

\section{Clinician Study: Quality of Rationales}

\subsection{Criteria for Clinical Reasoning Rationales}
\label{ssec:criteria}
To investigate the LLM's role as a clinical reasoner, assessing if the quality of clinical rationales meets the standards of clinicians, is of significant importance.
Prior works have incorporated human evaluations to assess the quality of machine-generated rationales~\citep{zelikman2022star, wang2023scott}. However, those works are limited to general domains, \ie, commonsense reasoning, and mainly focused on whether the rationales justify the target.

In this work, we and a group of licensed radiologists propose a novel set of criteria specifically designed to evaluate machine-generated rationales for clinical diagnosis. We expect this to facilitate and benefit future research on eliciting rationales for clinical application.
\begin{itemize}
  \item \textbf{\textsc{Consistency}}: How much a generated rationale is not contradictory to the presented data and model prediction (or the ground-truth diagnosis).
  \item \textbf{\textsc{Correctness}}: How medically correct the knowledge referred to in the rationale is.
  \item \textbf{\textsc{Specificity}}: How detailed and specific the insights provided in the generated rationale are.  
  \item \textbf{\textsc{Helpfulness}}: How much a clinical rationale benefits the prediction towards the correct diagnosis.
  \item \textbf{\textsc{Human-likeness}}: How well a clinical rationale demonstrates the insight and understanding of the presented patient description or diagnosis in a way that matches the human behaviours. 
\end{itemize}

\subsection{Human Evaluation}
\label{sec:human_intro}
We conduct human evaluation on 240 clinical reasoning rationales that are ``generated during the diagnosis'' with two radiologists. The results are illustrated in Figure~\ref{fig:human_eval}.
To assess them more critically, we sample rationales from ``challenging cases'' where ``all 5 models'' fail to predict the correct diagnosis (referred to as \textit{``Misdiagnoses''}), and an equal amount of correctly diagnosed cases (referred to as \textit{``Correct Diagnoses''}).
As a reference point, we apply the rationalizing module (Module I) to generate their rationales with access to ground-truth diagnosis (denoted as \textit{Ref}; gray-bar).

\paragraph{GPT-4 faithfully reflects available clinical evidence.}
We observe that \textit{GPT-4} and \textit{Ref} perform similarly within correct diagnoses. 
However, in \textit{Misdiagnoses} group, rationales generated without access to ground-truth diagnosis (\textit{GPT-4}) yield better scores even than those with access to it (\ie, \textit{Ref}), in every criterion. 
This phenomenon indicates that when a ground-truth diagnosis is challenging to predict based on the given patient description (\ie, \textit{Misdiagnosis} group), it is possible that prior knowledge of the diagnosis may not be beneficial for rationale annotation.
We presume it stems from the discrepancy of available clinical evidence between radiologists (who annotated the dataset in the real world) and our study. As a result, when asked to condition on the ground-truth label, GPT-4 may operate as if it is being forced to construct a reasoning path that contradicts its understanding of the available clinical evidence.
Therefore, the superior results of rationales generated without referencing to the ground truth in the \textit{Misdiagnoses} group (\textit{GPT-4}; yellow bars) manefest that GPT-4 can faithfully reflect the observed clinical evidence in the rationales.

\paragraph{Knowledge distillation enables better generalization.}
Intriguingly, the distilled student models also surpass \textit{Ref} in \textit{Misdiagnoses} cases.
This implies that training with clinical rationales helps the student model to generate a more generalized rationale that is not biased toward a certain diagnosis. Their better diagnostic performance than the LLM teacher supports this finding (Table~\ref{tab:student}).

\paragraph{Our framework elicits effective rationales for real-world applications.}
Overall, rationales generated by the teacher LLM (GPT-4) and student models receive scores higher than 4 regarding almost every criterion (the lowest is 3.45 with the score range being 0-5). 
The promising helpfulness and correctness scores in the misdiagnoses group match our previous findings: \textit{Misdiagnoses may not necessarily stem from ineffective rationales}, but rather from how we model the condition of rationales in diagnosis.
Also, high specificity scores indicate that both the LLM and distilled students can present detailed rationales concluding their observations and insights (the average length of the clinical rationale in this study is 269.4 words). 

Most importantly, the outstanding human-likeness scores, especially within the correct diagnoses, show that our rationales can effectively replicate the clinical reasoning of radiologists, and thus are more likely to be seamlessly integrated into real-world radiology reports.

\subsection{Case Study of Machine-generated Rationales}
We highlight two important properties of the generated rationales. 
We describe how they are aligned with the reasoning of radiologists, present the quotes from them, and elaborate on how they can benefit DL-based diagnosis.
\paragraph{Interpret clinical evidence contextually.}
The presence of the APOE4 gene is a strong risk factor for susceptibility to Alzheimer's disease. However, in situations where other evidence suggests that the patient is normal, radiologists do not just blindly rely on APOE4, instead, they comprehensively consider all the evidence based on their expertise and experiences. Our rationales exhibit the same behavior:
\begin{quote}
\textit{``The patient carries one copy of APOE4 gene, which is known to increase the risk of AD. The absence of cognitive impairment symptoms and brain atrophy \textbf{suggest that this genetic risk has not led to any apparent neurodegeneration}.''}
\end{quote}
In conventional data-driven approaches for disease diagnosis, \ie, text or image classification, it is non-trivial to annotate whether a certain feature is important in all possible scenarios. Our reasoning-aware diagnosis framework not only ameliorates this need, but also provides rationales that replicate such mechanism of human radiologists. Hence, we presume our framework has the potential to serve as a reliable tool for assisting clinicians in real-world data annotation.

\paragraph{Selectively summarize important evidence.}
In practice, after a thorough understanding and interpretation of all clinical evidence, radiologists select meaningful findings from their observations to make an accurate judgement.
Such summarization of evidence can be found in our rationales:
\begin{quote}
\textit{``... in summary, the patient's cognitive decline, as evidenced by the MMSE score and the presence of mild hippocampal and severe atrophy in key areas related to memory, indicate a mild cognitive impairment.''}
\end{quote}
Machine-generated rationales are often too long because they contain several sentences presenting the necessary information.
\citet{liu2023lost} demonstrate that when LLMs are processing long text, information at the beginning or the end of the text can be more effectively utilized than those scattered around. Since today's LLMs continue growing their ability to process longer texts, being able to selectively summarise necessary information at the end of the rationales supports our framework's usefulness for future LLM-based studies in clinical domains. 

\section{Related Work}
\paragraph{Alzheimer's Disease Diagnosis.}
Most DL-based methods for AD formulate the diagnosis simply as image classification and address the performance via transfer learning from general domains or tuning model architectures~\citep{ebrahimi2020introducing, jang2022m3t}. These approaches focus on better extracting the image features. However, AD diagnosis requires understanding and reasoning over a range of clinical data, such as APOE4 allele and MMSE alongside the MRIs~\citep{budson2012new, weller2018current}. To resolve that,
several studies have exploited different aspects or features of AD: \citet{zhu2016canonical} and \citet{zhang2018multi} approach AD diagnosis with multimodal data, \eg, positron emission
tomography and genetics data; \citet{ong2023evidence} leverage volume measurements of brain regions (\eg, subcortical volume) extracted from MRIs as additional training objectives alongside the classification of AD via multi-task learning. Although these studies do facilitate diagnosis models to consider more aspects or features of the disease, none of them provides a clear picture of the reasoning behind the diagnosis.

\paragraph{Clinical NLP.}
The success of LMs has sparked a surge in applying NLP techniques to the biomedical field~\citep{lee2020biobert, yue2020clinical, rajagopal2021cross, kim2023natural, feng2023chard}.
For example, \citet{lee2020biobert} fine-tune the commonly used BERT model~\citep{kenton2019bert} with medical corpus to endow it with biomedical knowledge, which is then implemented by~\citet{yue2020clinical} to solve the clinical reading comprehension task; \citet{rajagopal2021cross} and \citet{feng2023chard} address the generation of explanations for various medical conditions via sequence-to-sequence language models with template-based approaches.
More recently, upon the advancements of LLMs, \citet{agrawal2022large} have proposed to use LLMs to recognize named entities from clinical texts; 
\citet{li2023cancergpt} use LLMs to predict the synergy of drug pairs in rare human tissues that lack structured data and features.

Most prior work heavily relies on the knowledge from the pre-training corpus, ignoring whether the knowledge used is correct for the situation or follows human reasoning.
In this work, we address the absence of clinical reasoning, especially in disease diagnosis, via large language models with prompt-based learning.

\section{Conclusion}
We present a reasoning-aware diagnosis framework to target the absence of clinical reasoning in most prior works. Upon that, we investigate LLMs' reasoning ability in clinical diagnosis via prompt-based learning and embark on various experiments with few-shot diagnosis and knowledge distillation. 
As a formal study of clinical reasoning towards real-world applications, we propose a series of novel criteria for assessing the quality of machine-generated clinical rationales. These criteria can facilitate and benefit future work in this area. 
Through human evaluation and extensive analysis of generated rationales, we establish a solid foundation for utilizing LLMs, both directly and indirectly, to model clinical reasoning in disease diagnosis.

\paragraph{Limitations.}
Our study has the following limitations: (1) Our prompt used to invoke LLMs' CoT reasoning only contains two rationale demonstrations due to their length (around 260 words each). This can potentially affect models' performance in rationale generation and diagnosis; (2) In our settings, the clinical rationale and the name of the predicted diagnosis are autoregressively generated. We do not explore other paradigms, such as jointly predicting them via multi-task learning or dividing the rationale generation and diagnosis into separate stages; (3) Although a group of licensed radiologists are involved in this study, we have not incorporated this framework into real-world clinical settings; (4) We do not include filtering mechanism to target ineffective rationales. Although our analysis shows that even in misdiagnosed cases, 75\% of the rationales are medically correct, training student models with such a collection of rationales can potentially hinder their performance. 

\section{Ethical Statement}
All of the patient data from both ADNI and AIBL are approved by the institutional review boards and de-identified for privacy. Additionally, patient information in all examples provided in this paper is partially masked manually. Assessment for potential societal impacts regarding data bias, accountability, legal challenges, and so on, is necessary before applying our method to real clinical settings.

\section{Acknowledgements}
This work is mainly supported by the Samsung Research Funding Center of Samsung Electronics (Project Number SRFC-TF2103-01), and partially supported by Institute of Information \& Communications Technology Planning \& Evaluation (IITP) grant funded by the Korean government (MSIT) (No. 2020-0-01361, Artificial Intelligence Graduate School Program (Yonsei University)) and (No.2021-0-02068, Artificial Intelligence Innovation Hub) and (No.2022-0-00077, AI Technology Development for Commonsense Extraction, Reasoning, and Inference from Heterogeneous Data). Jinyoung Yeo is a corresponding author (jinyeo@yonsei.ac.kr).

\bibliography{aaai24}

\begin{thebibliography}{47}
\providecommand{\natexlab}[1]{#1}

\bibitem[{Agrawal et~al.(2022)Agrawal, Hegselmann, Lang, Kim, and
  Sontag}]{agrawal2022large}
Agrawal, M.; Hegselmann, S.; Lang, H.; Kim, Y.; and Sontag, D. 2022.
\newblock Large language models are few-shot clinical information extractors.
\newblock In \emph{Proceedings of the 2022 Conference on Empirical Methods in
  Natural Language Processing}, 1998--2022.

\bibitem[{Bakator and Radosav(2018)}]{bakator2018deep}
Bakator, M.; and Radosav, D. 2018.
\newblock Deep learning and medical diagnosis: A review of literature.
\newblock \emph{Multimodal Technologies and Interaction}, 2(3): 47.

\bibitem[{Balogh, Miller, and Ball(2015)}]{balogh2015improving}
Balogh, E.~P.; Miller, B.~T.; and Ball, J.~R. 2015.
\newblock Improving diagnosis in health care.

\bibitem[{Budson and Solomon(2012)}]{budson2012new}
Budson, A.~E.; and Solomon, P.~R. 2012.
\newblock New Criteria for Alzheimer’s disease and Mild Cognitive Impairment:
  Implications for the Practicing Clinician.
\newblock \emph{The neurologist}, 18(6): 356.

\bibitem[{Cook, Sherbino, and Durning(2018)}]{cook2018management}
Cook, D.~A.; Sherbino, J.; and Durning, S.~J. 2018.
\newblock Management reasoning: beyond the diagnosis.
\newblock \emph{Jama}, 319(22): 2267--2268.

\bibitem[{DeTure and Dickson(2019)}]{deture2019neuropathological}
DeTure, M.~A.; and Dickson, D.~W. 2019.
\newblock The neuropathological diagnosis of Alzheimer’s disease.
\newblock \emph{Molecular neurodegeneration}, 14(1): 1--18.

\bibitem[{Ebrahimi, Luo, and Chiong(2020)}]{ebrahimi2020introducing}
Ebrahimi, A.; Luo, S.; and Chiong, R. 2020.
\newblock Introducing transfer learning to 3D ResNet-18 for Alzheimer’s
  disease detection on MRI images.
\newblock In \emph{2020 35th international conference on image and vision
  computing New Zealand (IVCNZ)}, 1--6. IEEE.

\bibitem[{Ellis et~al.(2009)Ellis, Bush, Darby, De~Fazio, Foster, Hudson,
  Lautenschlager, Lenzo, Martins, Maruff et~al.}]{ellis2009australian}
Ellis, K.~A.; Bush, A.~I.; Darby, D.; De~Fazio, D.; Foster, J.; Hudson, P.;
  Lautenschlager, N.~T.; Lenzo, N.; Martins, R.~N.; Maruff, P.; et~al. 2009.
\newblock The Australian Imaging, Biomarkers and Lifestyle (AIBL) study of
  aging: methodology and baseline characteristics of 1112 individuals recruited
  for a longitudinal study of Alzheimer's disease.
\newblock \emph{International psychogeriatrics}, 21(4): 672--687.

\bibitem[{Feng et~al.(2023)Feng, Khetan, Sacaleanu, Gershman, and
  Hovy}]{feng2023chard}
Feng, S.~Y.; Khetan, V.; Sacaleanu, B.; Gershman, A.; and Hovy, E. 2023.
\newblock CHARD: Clinical Health-Aware Reasoning Across Dimensions for Text
  Generation Models.
\newblock In \emph{Proceedings of the 17th Conference of the European Chapter
  of the Association for Computational Linguistics}, 313--327.

\bibitem[{Hara, Kataoka, and Satoh(2017)}]{hara2017learning}
Hara, K.; Kataoka, H.; and Satoh, Y. 2017.
\newblock Learning spatio-temporal features with 3d residual networks for
  action recognition.
\newblock In \emph{Proceedings of the IEEE international conference on computer
  vision workshops}, 3154--3160.

\bibitem[{Hinton, Vinyals, and Dean(2015)}]{hinton2015distilling}
Hinton, G.; Vinyals, O.; and Dean, J. 2015.
\newblock Distilling the Knowledge in a Neural Network.
\newblock arXiv:1503.02531.

\bibitem[{Holzinger et~al.(2017)Holzinger, Biemann, Pattichis, and
  Kell}]{holzinger2017we}
Holzinger, A.; Biemann, C.; Pattichis, C.~S.; and Kell, D.~B. 2017.
\newblock What do we need to build explainable AI systems for the medical
  domain?
\newblock arXiv:1712.09923.

\bibitem[{Hsieh et~al.(2023)Hsieh, Li, Yeh, Nakhost, Fujii, Ratner, Krishna,
  Lee, and Pfister}]{hsieh-etal-2023-distilling}
Hsieh, C.-Y.; Li, C.-L.; Yeh, C.-k.; Nakhost, H.; Fujii, Y.; Ratner, A.;
  Krishna, R.; Lee, C.-Y.; and Pfister, T. 2023.
\newblock Distilling Step-by-Step! Outperforming Larger Language Models with
  Less Training Data and Smaller Model Sizes.
\newblock In \emph{Findings of the Association for Computational Linguistics:
  ACL 2023}, 8003--8017. Toronto, Canada: Association for Computational
  Linguistics.

\bibitem[{Jack~Jr et~al.(2008)Jack~Jr, Bernstein, Fox, Thompson, Alexander,
  Harvey, Borowski, Britson, L.~Whitwell, Ward et~al.}]{jack2008alzheimer}
Jack~Jr, C.~R.; Bernstein, M.~A.; Fox, N.~C.; Thompson, P.; Alexander, G.;
  Harvey, D.; Borowski, B.; Britson, P.~J.; L.~Whitwell, J.; Ward, C.; et~al.
  2008.
\newblock The Alzheimer's disease neuroimaging initiative (ADNI): MRI methods.
\newblock \emph{Journal of Magnetic Resonance Imaging: An Official Journal of
  the International Society for Magnetic Resonance in Medicine}, 27(4):
  685--691.

\bibitem[{Jang and Hwang(2022)}]{jang2022m3t}
Jang, J.; and Hwang, D. 2022.
\newblock M3T: three-dimensional Medical image classifier using Multi-plane and
  Multi-slice Transformer.
\newblock In \emph{Proceedings of the IEEE/CVF conference on computer vision
  and pattern recognition}, 20718--20729.

\bibitem[{Kassirer(1989)}]{kassirer1989diagnostic}
Kassirer, J.~P. 1989.
\newblock Diagnostic reasoning.
\newblock \emph{Annals of internal medicine}, 110(11): 893--900.

\bibitem[{Kenton and Toutanova(2019)}]{kenton2019bert}
Kenton, J. D. M.-W.~C.; and Toutanova, L.~K. 2019.
\newblock Bert: Pre-training of deep bidirectional transformers for language
  understanding.
\newblock In \emph{Proceedings of NAACL-HLT}, volume~1, 2.

\bibitem[{Kim et~al.(2023)Kim, Ong, Choi, Yeo, Kim, Han, Park, Kim, Choi, Ahn
  et~al.}]{kim2023natural}
Kim, M.; Ong, K. T.-i.; Choi, S.; Yeo, J.; Kim, S.; Han, K.; Park, J.~E.; Kim,
  H.~S.; Choi, Y.~S.; Ahn, S.~S.; et~al. 2023.
\newblock Natural language processing to predict isocitrate dehydrogenase
  genotype in diffuse glioma using MR radiology reports.
\newblock \emph{European Radiology}, 33(11): 8017--8025.

\bibitem[{Kingma and Ba(2014)}]{kingma2014adam}
Kingma, D.~P.; and Ba, J. 2014.
\newblock Adam: A method for stochastic optimization.
\newblock \emph{arXiv preprint arXiv:1412.6980}.

\bibitem[{Kojima et~al.(2022)Kojima, Gu, Reid, Matsuo, and
  Iwasawa}]{kojima2022large}
Kojima, T.; Gu, S.~S.; Reid, M.; Matsuo, Y.; and Iwasawa, Y. 2022.
\newblock Large language models are zero-shot reasoners.
\newblock \emph{Advances in neural information processing systems}, 35:
  22199--22213.

\bibitem[{Kumar et~al.(2022)Kumar, Koul, Singla, and
  Ijaz}]{kumar2022artificial}
Kumar, Y.; Koul, A.; Singla, R.; and Ijaz, M.~F. 2022.
\newblock Artificial intelligence in disease diagnosis: a systematic literature
  review, synthesizing framework and future research agenda.
\newblock \emph{Journal of ambient intelligence and humanized computing},
  1--28.

\bibitem[{Lee et~al.(2020)Lee, Yoon, Kim, Kim, Kim, So, and
  Kang}]{lee2020biobert}
Lee, J.; Yoon, W.; Kim, S.; Kim, D.; Kim, S.; So, C.~H.; and Kang, J. 2020.
\newblock BioBERT: a pre-trained biomedical language representation model for
  biomedical text mining.
\newblock \emph{Bioinformatics}, 36(4): 1234--1240.

\bibitem[{Li et~al.(2023)Li, Shetty, Kamath, Jaiswal, Jiang, Ding, and
  Kim}]{li2023cancergpt}
Li, T.; Shetty, S.; Kamath, A.; Jaiswal, A.; Jiang, X.; Ding, Y.; and Kim, Y.
  2023.
\newblock CancerGPT: Few-shot Drug Pair Synergy Prediction using Large
  Pre-trained Language Models.
\newblock arXiv:2304.10946.

\bibitem[{Li{\'e}vin, Hother, and Winther(2023)}]{lievin2022can}
Li{\'e}vin, V.; Hother, C.~E.; and Winther, O. 2023.
\newblock Can large language models reason about medical questions?
\newblock arXiv:2207.08143.

\bibitem[{Liu et~al.(2023)Liu, Lin, Hewitt, Paranjape, Bevilacqua, Petroni, and
  Liang}]{liu2023lost}
Liu, N.~F.; Lin, K.; Hewitt, J.; Paranjape, A.; Bevilacqua, M.; Petroni, F.;
  and Liang, P. 2023.
\newblock Lost in the Middle: How Language Models Use Long Contexts.
\newblock arXiv:2307.03172.

\bibitem[{Mitra et~al.(2020)Mitra, Banerjee, Pal, Mishra, and
  Baral}]{mitra2019exploring}
Mitra, A.; Banerjee, P.; Pal, K.~K.; Mishra, S.; and Baral, C. 2020.
\newblock How Additional Knowledge can Improve Natural Language Commonsense
  Question Answering?
\newblock arXiv:1909.08855.

\bibitem[{Norman(2005)}]{norman2005research}
Norman, G. 2005.
\newblock Research in clinical reasoning: past history and current trends.
\newblock \emph{Medical education}, 39(4): 418--427.

\bibitem[{Ong et~al.(2023)Ong, Kim, Kim, Jang, Sohn, Choi, Hwang, Hwang, and
  Yeo}]{ong2023evidence}
Ong, K. T.-i.; Kim, H.; Kim, M.; Jang, J.; Sohn, B.; Choi, Y.~S.; Hwang, D.;
  Hwang, S.~J.; and Yeo, J. 2023.
\newblock Evidence-empowered Transfer Learning for Alzheimer's Disease.
\newblock arXiv:2303.01105.

\bibitem[{OpenAI(2023{\natexlab{a}})}]{openai2023chatgpt}
OpenAI. 2023{\natexlab{a}}.
\newblock ChatGPT.
\newblock \url{https://openai.com/blog/chatgpt}.

\bibitem[{OpenAI(2023{\natexlab{b}})}]{openai2023gpt4}
OpenAI. 2023{\natexlab{b}}.
\newblock GPT-4 Technical Report.

\bibitem[{Qiu et~al.(2020)Qiu, Joshi, Miller, Xue, Zhou, Karjadi, Chang, Joshi,
  Dwyer, Zhu et~al.}]{qiu2020development}
Qiu, S.; Joshi, P.~S.; Miller, M.~I.; Xue, C.; Zhou, X.; Karjadi, C.; Chang,
  G.~H.; Joshi, A.~S.; Dwyer, B.; Zhu, S.; et~al. 2020.
\newblock Development and validation of an interpretable deep learning
  framework for Alzheimer’s disease classification.
\newblock \emph{Brain}, 143(6): 1920--1933.

\bibitem[{Rajagopal et~al.(2021)Rajagopal, Khetan, Sacaleanu, Gershman, Fano,
  and Hovy}]{rajagopal2021cross}
Rajagopal, D.; Khetan, V.; Sacaleanu, B.; Gershman, A.; Fano, A.; and Hovy, E.
  2021.
\newblock Cross-domain reasoning via template filling.
\newblock \emph{arXiv preprint arXiv:2111.00539}.

\bibitem[{Touvron et~al.(2023)Touvron, Martin, Stone, Albert, Almahairi,
  Babaei, Bashlykov, Batra, Bhargava, Bhosale et~al.}]{touvron2023llama2}
Touvron, H.; Martin, L.; Stone, K.; Albert, P.; Almahairi, A.; Babaei, Y.;
  Bashlykov, N.; Batra, S.; Bhargava, P.; Bhosale, S.; et~al. 2023.
\newblock Llama 2: Open Foundation and Fine-Tuned Chat Models.
\newblock arXiv:2307.09288.

\bibitem[{Tsimpoukelli et~al.(2021)Tsimpoukelli, Menick, Cabi, Eslami, Vinyals,
  and Hill}]{tsimpoukelli2021multimodal}
Tsimpoukelli, M.; Menick, J.~L.; Cabi, S.; Eslami, S.; Vinyals, O.; and Hill,
  F. 2021.
\newblock Multimodal few-shot learning with frozen language models.
\newblock \emph{Advances in Neural Information Processing Systems}, 34:
  200--212.

\bibitem[{Voss et~al.(2017)Voss, Bridge, Cohen, and Walker}]{voss2017closer}
Voss, J.~L.; Bridge, D.~J.; Cohen, N.~J.; and Walker, J.~A. 2017.
\newblock A closer look at the hippocampus and memory.
\newblock \emph{Trends in cognitive sciences}, 21(8): 577--588.

\bibitem[{Wang et~al.(2023)Wang, Wang, Li, Gao, Yin, and Ren}]{wang2023scott}
Wang, P.; Wang, Z.; Li, Z.; Gao, Y.; Yin, B.; and Ren, X. 2023.
\newblock {SCOTT}: Self-Consistent Chain-of-Thought Distillation.
\newblock In \emph{Proceedings of the 61st Annual Meeting of the Association
  for Computational Linguistics (Volume 1: Long Papers)}, 5546--5558. Toronto,
  Canada: Association for Computational Linguistics.

\bibitem[{Wei et~al.(2022)Wei, Wang, Schuurmans, Bosma, Xia, Chi, Le, Zhou
  et~al.}]{wei2022chain}
Wei, J.; Wang, X.; Schuurmans, D.; Bosma, M.; Xia, F.; Chi, E.; Le, Q.~V.;
  Zhou, D.; et~al. 2022.
\newblock Chain-of-thought prompting elicits reasoning in large language
  models.
\newblock \emph{Advances in Neural Information Processing Systems}, 35:
  24824--24837.

\bibitem[{Weller and Budson(2018)}]{weller2018current}
Weller, J.; and Budson, A. 2018.
\newblock Current understanding of Alzheimer’s disease diagnosis and
  treatment.
\newblock \emph{F1000Research}, 7.

\bibitem[{Wu, Zhang, and Huang(2023)}]{wu2023chain}
Wu, D.; Zhang, J.; and Huang, X. 2023.
\newblock Chain of Thought Prompting Elicits Knowledge Augmentation.
\newblock In \emph{Findings of the Association for Computational Linguistics:
  ACL 2023}, 6519--6534. Toronto, Canada: Association for Computational
  Linguistics.

\bibitem[{Yue, Jimenez~Gutierrez, and Sun(2020)}]{yue2020clinical}
Yue, X.; Jimenez~Gutierrez, B.; and Sun, H. 2020.
\newblock Clinical Reading Comprehension: A Thorough Analysis of the emr{QA}
  Dataset.
\newblock In \emph{Proceedings of the 58th Annual Meeting of the Association
  for Computational Linguistics}, 4474--4486. Online: Association for
  Computational Linguistics.

\bibitem[{Zelikman et~al.(2022)Zelikman, Wu, Mu, and
  Goodman}]{zelikman2022star}
Zelikman, E.; Wu, Y.; Mu, J.; and Goodman, N. 2022.
\newblock Star: Bootstrapping reasoning with reasoning.
\newblock \emph{Advances in Neural Information Processing Systems}, 35:
  15476--15488.

\bibitem[{Zhang et~al.(2018)Zhang, Adeli, Zhou, Chen, and
  Shen}]{zhang2018multi}
Zhang, C.; Adeli, E.; Zhou, T.; Chen, X.; and Shen, D. 2018.
\newblock Multi-Layer Multi-View Classification for Alzheimer's Disease
  Diagnosis.
\newblock In \emph{Proceedings of the Thirty-Second AAAI Conference on
  Artificial Intelligence and Thirtieth Innovative Applications of Artificial
  Intelligence Conference and Eighth AAAI Symposium on Educational Advances in
  Artificial Intelligence}, AAAI'18/IAAI'18/EAAI'18. AAAI Press.
\newblock ISBN 978-1-57735-800-8.

\bibitem[{Zhang et~al.(2022)Zhang, Roller, Goyal, Artetxe, Chen, Chen, Dewan,
  Diab, Li, Lin, Mihaylov, Ott, Shleifer, Shuster, Simig, Koura, Sridhar, Wang,
  and Zettlemoyer}]{zhang2022opt}
Zhang, S.; Roller, S.; Goyal, N.; Artetxe, M.; Chen, M.; Chen, S.; Dewan, C.;
  Diab, M.; Li, X.; Lin, X.~V.; Mihaylov, T.; Ott, M.; Shleifer, S.; Shuster,
  K.; Simig, D.; Koura, P.~S.; Sridhar, A.; Wang, T.; and Zettlemoyer, L. 2022.
\newblock OPT: Open Pre-trained Transformer Language Models.
\newblock arXiv:2205.01068.

\bibitem[{Zhang et~al.(2023)Zhang, Zhang, Li, Zhao, Karypis, and
  Smola}]{zhang2023multimodal}
Zhang, Z.; Zhang, A.; Li, M.; Zhao, H.; Karypis, G.; and Smola, A. 2023.
\newblock Multimodal Chain-of-Thought Reasoning in Language Models.
\newblock arXiv:2302.00923.

\bibitem[{Zheng et~al.(2022)Zheng, Li, Zhang, Zhuang, Chen, Huang, Wang, Xu,
  Zhuo, Xing et~al.}]{zheng2022alpa}
Zheng, L.; Li, Z.; Zhang, H.; Zhuang, Y.; Chen, Z.; Huang, Y.; Wang, Y.; Xu,
  Y.; Zhuo, D.; Xing, E.~P.; et~al. 2022.
\newblock Alpa: Automating inter- and intra-operator parallelism for
  distributed deep learning.
\newblock In \emph{16th USENIX Symposium on Operating Systems Design and
  Implementation (OSDI 22)}, 559--578.

\bibitem[{Zhu et~al.(2021)Zhu, Sun, Huang, Han, and Zhang}]{zhu2021dual}
Zhu, W.; Sun, L.; Huang, J.; Han, L.; and Zhang, D. 2021.
\newblock Dual attention multi-instance deep learning for Alzheimer’s disease
  diagnosis with structural MRI.
\newblock \emph{IEEE Transactions on Medical Imaging}, 40(9): 2354--2366.

\bibitem[{Zhu et~al.(2016)Zhu, Suk, Lee, and Shen}]{zhu2016canonical}
Zhu, X.; Suk, H.-I.; Lee, S.-W.; and Shen, D. 2016.
\newblock Canonical feature selection for joint regression and multi-class
  identification in Alzheimer’s disease diagnosis.
\newblock \emph{Brain imaging and behavior}, 10: 818--828.

\end{thebibliography}

\clearpage
\section{Appendix}
\setcounter{secnumdepth}{1}
\section{Details on the Datasets}
We list details on the datasets we used in this study, as well as the pre-processing procedure we applied to transfer the MRI scan.
We collect data for Alzheimer's disease (AD) diagnosis from two organizations/projects.
 
\subsubsection{ADNI}
The Alzheimer’s Disease Neuroimaging Initiative (ADNI)~\citep{jack2008alzheimer} is a long-term project/study on AD. ADNI brings together researchers with study data as they work to define the progression and address the diagnosis of AD. Data from ADNI has been widely used in prior works and profoundly influenced the development of DL-based AD diagnosis~\citep{ebrahimi2020introducing,zhang2018multi, jang2022m3t, ong2023evidence}.

\subsubsection{AIBL}
The Australian Imaging, Biomarker and Lifestyle Flagship Study of Ageing (AIBL)~\citep{ellis2009australian} is a project to investigate which biomarkers, cognitive characteristics, and health and lifestyle factors determine the subsequent progression of symptomatic AD. Data from AIBL is also one of the most widely used data for AD diagnosis~\citep{qiu2020development, zhu2021dual, jang2022m3t}.

\subsection{Common Features of Both Datasets}
Each data we collect from ADNI and AIBL has the following components: (1) MRI scans; (2) demographic information; (3) education level; (4) results from the mini-mental state examination; (5) the presence of APOE4 allele; (6) The ground-truth label of diagnosis.

\subsection{Transforming MRIs into Descriptions} 
The MRI scans belong to the imaging modality. However, LLMs only accept textual inputs unless tailored adjustments are made. Hence, we incorporate an automatic process to transform the structural features of brain regions\footnote{We select 14 regions that are associated with AD: Hippocampus, Amygdala, Entorhinal, Parahippocampus, Medial Temporal Lobe, Fusiform, Precuneus, Superior Paretal, Lateral Ventricle, Frontal Lobe, Temporal Lobe, Parietal Lobe, Occipital Lobe, and Cerebral Cortex.}

We first sort patient cases into several groups based on their demographic information and extract the volume measurements of brain regions from patients' MRI scans (\eg, cortical thickness measurements). After that, within each group, we use the average volume of NC,
MCI, and AD cases to define the
interval for data annotation (\eg, The average subcortical volume of all AD cases in group 1). 
If a region’s volume is larger than the mean of
NC cases’ average and MCI cases’ average, we label its level
of atrophy as \textbf{\textit{NO}}. If a region’s volume is smaller than the
mean of AD cases’ average and MCI cases’ average, we label
its atrophy level as \textbf{\textit{SEVERE}}. Those in between are labelled as
\textbf{\textit{MILD}}.

The textualized MRI features are combined with components (2) - (5) mentioned above to form a complete patient description $\mathcal{P}$ of a patient case.

\subsection{Statistics}
We acquire 7,124 data for Alzheimer's disease diagnosis from ADNI and 428 from AIBL.
Data from ADNI are split into training, validation, and test sets.
All the AIBL data are exclusively used as an out-of-domain test set rather than training.
The statistics is provided in Table~\ref{tab:dataset_statistics}

\subsection{Additional Statement on the Datasets}
All data used in this work are approved by the Institutional
Review Board. They should not be shared without permission and
only be used by authorized researchers for research purposes. Therefore, in the following section: (1) \textbf{We do not provide any MRI scans}; (2) The patient information presented in the prompt is \textbf{partially masked or omitted}; (3) We \textbf{only present one demonstration} in the prompt.

\begin{table}[t!]
\centering
\resizebox{0.9\columnwidth}{!}{
\begin{tabular}{l|ccc}
\toprule
Dataset & \# Train & \# Valid & \# Test \\ 
\midrule
ADNI & 6,062 & 303 & 759\\
AIBL & - & - & 428\\
\midrule
Within the test set & \# AD & \# MCI & \# NC \\
\midrule
ADNI & 248 & 259 & 252 \\
AIBL & 130 & 158 & 140 \\
\bottomrule
\end{tabular}%
}
\caption{Statistics of the two datasets.}
\label{tab:dataset_statistics}
\end{table}

\section{Prompts and Experimental Details}
\subsection{Prompts}
\paragraph{Generating rationale candidates.}
We first apply a zero-shot chain-of-thought (CoT) prompting to generate 20 candidate clinical rationales. This prompt is in Table~\ref{tab:appendix_prompt}. After that, a group of licensed radiologists select two clinical rationales as the CoT demonstrations used in the following prompts.

\paragraph{Prompt for the rationalization module (Module I).}
This prompt is based on patient description, CoT demonstration, and the ground-truth label of diagnosis (Table~\ref{tab:appendix_prompt}). We only present one CoT demonstration in this appendix.

\paragraph{Prompt for few-shot diagnosis with LLMs (Module II-1).}
This prompt is similar to the previous prompt but without offering the ground-truth diagnosis. Instead, we use the form of multi-choice questions to invoke LLMs' diagnosis. See Table~\ref{tab:appendix_prompt}.

\subsection{Implementation Details of Student Models} 
Experiments are run on 8 NVIDIA RTX A6000 GPUs, each with 48 GB of RAM. During training, we set the batch size to 8, the learning rate to $2\times10^{-3}$, and use the Adam optimizer~\citep{kingma2014adam}. We set the maximum number of epochs to 15 and select the checkpoint with the lowest validation loss.

\begin{table*}[t]
\centering
\resizebox{\textwidth}{!}{%
\begin{tabular}{l}
\toprule
\textbf{Prompt 1: Candidate Rationale Generation} \\
Generate detailed medical rationales for the diagnosis ("Diagnosis:") based on the patient description.\\These rationales should be the crucial cue for the diagnosis. Pretend that you don't know the diagnosis (``Diagnosis'').
\\\textbf{Patient Description:} \textbf{Diagnosis:} \textbf{Medical Rationale:} \\
\midrule
\textbf{Prompt 2: Clinical Rationalization} \\
Generate detailed medical rationales for the diagnosis ("Diagnosis:") based on the patient description.\\These rationales should be the crucial cue for the diagnosis. Pretend that you don't know the diagnosis ("Diagnosis:")\\\textbf{Example 1}\\\textbf{Patient Description:} This patient is a XX-year-old Male who has completed 16 years of education and is Married.\\ The patient has a Mini-mental State Examination score of 26.0/30 and has no APOE4 gene.\\Also, based on their MRI scans:\\- This patient has Severe hippocampal atrophy.\\- This patient has ...\\ ......\\\textbf{Diagnosis:}  Alzheimer's Disease\\\textbf{Medical Rationale:}\\The patient, a XX-year-old male presents with severe atrophy in multiple areas of the brain as determined by MRI scans.\\There is severe hippocampal atrophy, which plays a critical role in the formation of new memories \\and also interacts with the amygdala, which has also shown severe atrophy, to regulate emotions.\\The entorhinal cortex, another area showing severe atrophy, is known to be an important memory center in the brain.\\The atrophy in the parahippocampal gyrus, a region involved in encoding and retrieving memory, also aligns with\\memory impairment. Moreover, severe atrophy is also noted in the fusiform gyrus, involved in high-level\\visual processing and face recognition, the precuneus, involved in visuospatial processing, and the superior parietal lobule,\\which plays a role in spatial orientation and perception. Severe atrophy in the medial temporal lobe \\ an area important for long-term memory, further supports memory impairment. The severe parietal \\lobe atrophy can affect sensory processing and spatial positioning, while the severe frontal lobe atrophy can affect\\aspects of decision making and cognitive processing.The severe occipital lobe atrophy can impact visual processing.\\Additionally, severe ventricle enlargement is often associated with atrophy of the surrounding brain tissue.\\Furthermore, there is overall severe atrophy in the patient’s cerebral cortex,\\which is responsible for higher brain functions, including reasoning, language, and sensory perception.\\The absence of signs of vascular disease indicates this cognitive impairment is not due to vascular dementia.\\The Mini-mental State Examination score of 26.0/30 indicates mild cognitive impairment.\\Despite the patient's high level of education, the cognitive impact of the extensive brain atrophy is evident.\\Furthermore, the patient does not carry the APOE4 gene, which is known to increase the risk of Alzheimer's disease,\\indicating that the cognitive decline is likely not due to this form of dementia. \\Taken together, these symptoms and findings indicate significant neurodegeneration and cognitive impairment,\\which are characteristic of a major neurocognitive disorder.\\\textbf{Example 2} ......omitted......\\\textbf{Example 3}\\\textbf{Patient Description:}\\\textbf{Diagnosis:}\\\textbf{Medical Rationale:}\\
\midrule
\textbf{Prompt 3: Few-shot Diagnosis with LLMs} \\
You are a doctor. Generate medical rationale and diagnose the patient based on the information your colleague gave you.\\You can utilize the medical rationale. Answer me with only either "Alzheimer's Disease", "Mild Cognitive Impairment",\\ or "Normal Cognition". You should follow the style of the history.\\
\textbf{Example 1}\\\textbf{Patient Description:} ......same as Prompt 2......\\\textbf{What is the diagnosis of this patient?}\\(A) Alzheimer's Disease (B) Mild Cognitive Impairment (C) Normal Cognition
\\\textbf{Medical Rationale:} ......same as Prompt 2......\\\textbf{Diagnosis:} (A) Alzheimer's Disease\\\textbf{Example 2} ......omitted......\\\textbf{Example 3}\\\textbf{Patient Description:}\\
\bottomrule
\end{tabular}%
}
\caption{Prompts used in this study.}
\label{tab:appendix_prompt}
\end{table*}

\end{document}